\documentclass{article}
\usepackage[a4paper, total={6in, 9.5in}]{geometry}
\usepackage[dvipsnames]{xcolor}
\usepackage{mathtools} \usepackage{multirow}
\usepackage{booktabs}
\usepackage{colortbl}
\usepackage{subfigure}
\usepackage[export]{adjustbox}
\usepackage{afterpage}

\usepackage{wrapfig}

\usepackage{caption}

\usepackage[utf8]{inputenc} % allow utf-8 input
\usepackage[T1]{fontenc}    % use 8-bit T1 fonts
\usepackage{hyperref}       % hyperlinks
\usepackage{url}            % simple URL typesetting
\usepackage{booktabs}       % professional-quality tables
\usepackage{amsfonts}       % blackboard math symbols
\usepackage{nicefrac}       % compact symbols for 1/2, etc.
\usepackage{microtype}      % microtypography
\usepackage{xcolor}         % colors

\def\eg{\emph{e.g.~}}

\def\aka{\emph{a.k.a.~}}

\usepackage{tikz}

\title{Generated Faces in the Wild: Quantitative Comparison of Stable Diffusion, Midjourney and DALL-E 2}

\author{
Ali Borji \\
Quintic AI, San Francisco, CA \\ 
\texttt{aliborji@gmail.com} 
}

\begin{document}

\maketitle

\begin{abstract}
The field of image synthesis has made great strides in the last couple of years. Recent models are capable of generating images with astonishing quality. Fine-grained evaluation of these models on some interesting categories such as faces is still missing. Here, we conduct a quantitative comparison of three popular systems including Stable Diffusion, Midjourney, and DALL-E 2 in their ability to generate photorealistic faces in the wild. We find that Stable Diffusion generates better faces than the other systems, according to the FID score. We also introduce a dataset of generated faces in the wild dubbed GFW, including a total of 15,076 faces. Furthermore, we hope that our study spurs further research in assessing the generative models and improving them. Data and code are available at \href{https://drive.google.com/file/d/16BXO1fgN08UGLLeA5ZNU9bhwAkcAOdci/view?usp=sharing}{\underline{data}} and \href{https://github.com/aliborji/GFW}{\underline{code}}, respectively.

\end{abstract}

\section{Introduction}
The field of image synthesis has made great strides in the last couple of years. Variations of Generative Adversarial Networks (GAN)~\cite{goodfellow2020generative} and Variational Autoencoders (VAE)~\cite{kingma2013auto} were the first to generate high quality images. Recent diffusion-based models trained on massive datasets have transcended progress and have attracted a lot of attention both among AI scientists and the public. Several blog posts (\eg \href{https://spectrum.ieee.org/openai-dall-e-2}{here}, \href{https://towardsdatascience.com/dall-e-2-explained-the-promise-and-limitations-of-a-revolutionary-ai-3faf691be220}{here}, and \href{https://www.lesswrong.com/posts/uKp6tBFStnsvrot5t/what-dall-e-2-can-and-cannot-do}{here}) 
and articles (\eg~\cite{marcus2022very}) have offered anecdotal evaluations. Fine-grained evaluation of these models on some interesting categories still needs more work. 

Quantitative evaluation of generative models is mostly concerned with fidelity and diversity of the entire scenes rather than scene components or individual objects (with few exceptions~\cite{zhao2018bias,bau2019seeing}). Here, we emphasize on fine-grained evaluation of models, in particular their ability to generate photorealistic faces. Notice that our work is different from those studies that attempt to build models for generating portraits or evaluating such models (\eg StyleGAN~\cite{karras2020analyzing}). Instead, here we are interested in evaluating the quality of generated faces in cluttered scenes containing multiple objects. To this end, we utilize text to image generative models to synthesize scenes. We then use a face detector to detect faces in these images. Finally, we use the well-established Fr\'echet Inception Distance (FID)~\cite{heusel2017gans} to evaluate the quality of the generated faces against a set of real faces.

\section{Comparison}

\subsection{Models}

We consider the following three models:
\begin{enumerate}
    \item {\bf Stable Diffusion~\cite{rombach2022high}}\footnote{\url{https://en.wikipedia.org/wiki/Stable_Diffusion}}. Released by StabilityAI in 2022, this model is primarily used to generate detailed images conditioned on text descriptions. It can also be applied to other tasks such as inpainting, outpainting, and image translation. This model is trained on 512 $\times$ 512 images from a subset of the LAION-5B database. It uses a frozen CLIP ViT-L/14 text encoder to condition the model on text prompts. With its 860M UNet and 123M text encoder, the model is relatively lightweight and runs on a GPU with at least 10GB VRAM. We use the Colab notebook provided by Hugging Face\footnote{\url{https://colab.research.google.com/github/huggingface/notebooks/blob/main/diffusers/stable_diffusion.ipynb}} to run Stable Diffusion.

    \item {\bf Midjourney (\url{https://www.midjourney.com/})}\footnote{\url{https://en.wikipedia.org/wiki/Midjourney}}. This model was created by an independent research lab with the same name. It can synthesize images from textual descriptions and is currently in open beta. 
    Midjourney tends to generate surrealistic images and is popular among artists. We used a collection of images generated by this model available via this Kaggle  \href{https://www.kaggle.com/datasets/da9b9ba35ffbd86a5f97ccd068d3c74f5742cfe5f34f6aaf1f0f458d7694f55e?resource=download}{link}.

    \item {\bf DALL-E 2~\cite{ramesh2022hierarchical}}\footnote{\url{https://en.wikipedia.org/wiki/DALL-E}, \url{https://openai.com/dall-e-2/}} is created by OpenAI and is a successor of DALL-E. It can create more realistic images than DALL-E at higher resolutions and can combine concepts, attributes, and styles. DALL-E 2 is trained on approximately 650 million image-text pairs scraped from the Internet. Since DALL-E 2 code is not available, we were not able to generate images on a large scale. We accessed the system via their portal by manually entering the prompts and saving the results.

\end{enumerate}

\subsection{Data}
To evaluate the models, two sets of faces are required: a) generated faces, and b) real faces.

\begin{itemize}
    \item {\bf Generated faces.} We avoided crawling the web for generated faces to reduce the potential biases. Indeed, people usually tend to post high quality faces in social media (\aka cherry picking). Instead, we use the captions from the COCO dataset\footnote{\url{https://cocodataset.org/}} (captions\_train2017.json) as prompts to synthesize images. Images generated by Stable Diffusion and DALL-E 2 have size of 512 $\times$ 512, while images by Midjourney are of variable size. To increase the chance of generated images to contain faces, we chose captions that had any word from the following list of words: \texttt{[`person', `man', `woman', `men', `women', `kid', `child', `face', `girl', `boy']}. We then ran the MediaPipe face detector\footnote{\url{https://google.github.io/mediapipe/solutions/face_detection.html}} twice: first on the entire image to detect faces, and a second time on individual detections to prune false positives. Finally, we manually removed the remaining false positives. Detected faces were resized to 100 $\times$ 100 pixels. In total, we collected 15,076 generated faces, including 8,050 by Stable Diffusion, 6,350 by Midjourney, and 676 by DALL-E 2.

    \item {\bf Real faces.} We ran the face detector on the COCO training set (train2017.zip) similar to above to extract faces. In addition, we added 13,233 faces from the Labeled Faces in the Wild (LFW) \footnote{\url{http://vis-www.cs.umass.edu/lfw/}} dataset~\cite{huang2008labeled}. We cropped the central 100 $\times$ 100 pixels area from the 250 $\times$ 250 faces of LFW. In total, we collected 30,000 real faces.

We removed faces (both generated and real) that were highly occluded (\eg person eating a large piece of food, or fruit or wearing a mask) as well as too dark or too blurry faces. We also removed dull faces, animal faces, drawings, and cartoons. Faces with eyeglasses were kept. %We also tried to make the faces upright.

\end{itemize}

\subsection{Evaluation Scheme}
Several scores have been proposed for evaluating generative models~\cite{borji2019pros,borji2022pros}. The FID score is the most commonly used one. We utilize the implementation from 
\url{https://github.com/mseitzer/pytorch-fid}. Each time we shuffled the faces (generated and real) and randomly selected 5,000 faces from each set, and computed the FID score between the two sets. We repeated this procedure 10 times and computed the mean and standard deviation of FID across the runs. Since DALL-E 2 faces are less than 5K, we selected faces with replacement for this model. To compute the FID for real faces, we first shuffle the set of real faces and then split the first 10K faces into two sets of 5K faces. The FID between these two sets is then computed.

\subsection{Results}

\begin{figure}[t]       
    \centering 
    \includegraphics[width=.9\linewidth]{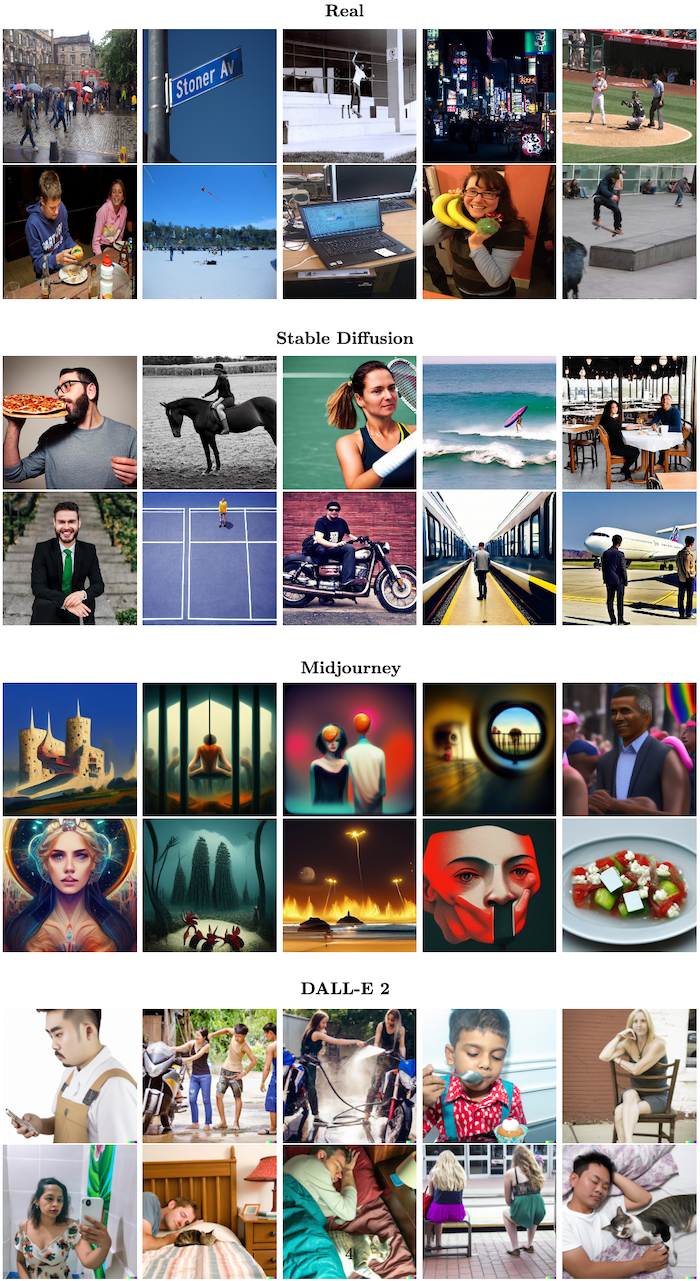}
    \caption{Samples of real images (top row) and generated images.}
    \label{fig:imgs}
\end{figure}

\begin{figure}[t]       
    \centering 
    \includegraphics[width=.9\linewidth]{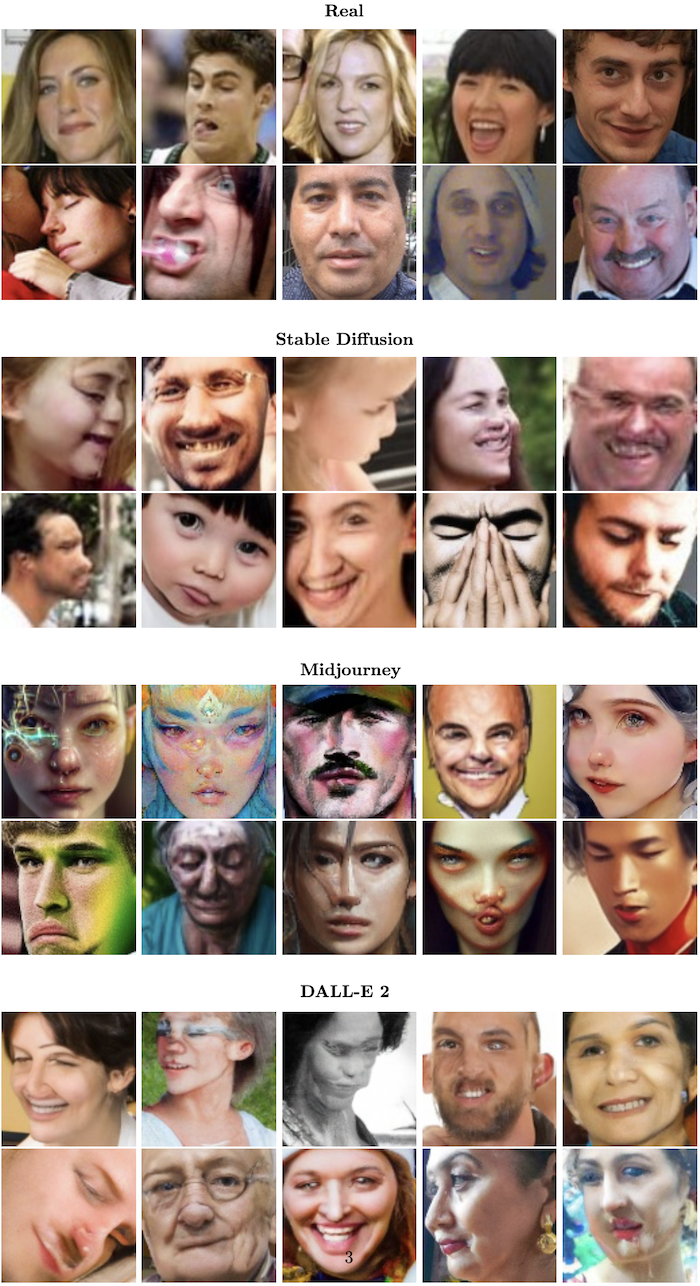}
    \caption{Samples of real faces (top row) and generated faces.}
    \label{fig:face}
\end{figure}

Sample images and faces generated by three systems are shown in Fig.~\ref{fig:imgs} and Fig.~\ref{fig:face}, respectively. Qualitative inspection shows that models are able to generate high quality images. The generated images have high fidelity, are diverse, and most of the time are physically plausible (with few exceptions \eg the right woman in the last row of Fig.~\ref{fig:imgs}). The generated faces (Fig.~\ref{fig:face}) in general look very good, although not as good as the faces generated by models specifically trained on portraits (\eg StyleGAN~\cite{karras2020analyzing}). 

\begin{figure}[t]       
    \centering 
    % \vspace{-10pt}
    \includegraphics[width=.47\linewidth]{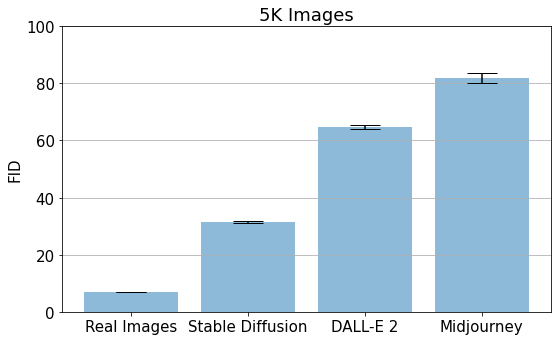}
    \includegraphics[width=.47\linewidth]{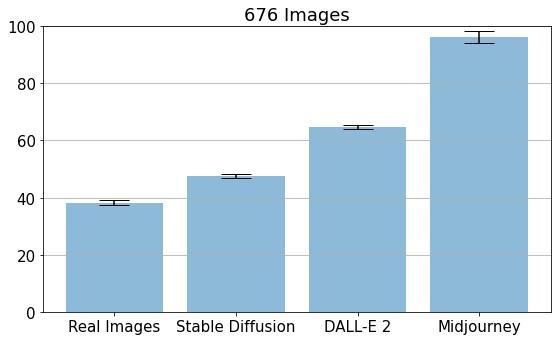}
    \caption{Left: FID scores of models over random sets of 5000 faces. Right: Results with samples of size 676 per model. Notice that the lower the FID, the better. Results are averaged over 10 runs.}
    \label{fig:res}
\end{figure}

FID scores are shown in Fig.~\ref{fig:res}. Stable Diffusion scores a lower FID, hence generating better faces than the other two models. The quality of the generated faces, however, is still much worse than the real faces, witnessed by the large gap in FID of models and real faces. The poor performance of Midjourney can be partly due to its generated faces being surrealistic and anime. 

There are three possible reasons why DALL-E 2 performs worse than Stable Diffusion. First, OpenAI introduced some deepfake safeguards during DALL-E 2 training to prevent it from memorizing faces that often appear on the Internet. Second, DALL-E 2 is optimized for images with a single focus of attention. Thus, this model does better at generating portraits\footnote{DALL-E 2 is able to generate realistic portraits. Please see \href{https://twitter.com/Merzmensch/status/1539366833920286722}{this twitter post} for some samples.} of imaginary people than faces in complex scenes. Third, lower performance can be attributed to the smaller set of images for this model compared to the other two. It is known that FID is sensitive to the number of images. The higher number of images, the more reliable the score. We address this issue in the following. 

To make a fair comparison of the systems, we limited the number of faces to 676 which is the number of DALL-E 2 faces. Results are shown in the right panel of Fig.~\ref{fig:res}. As can be seen, FID values are higher now, but still Stable Diffusion wins. We believe larger sets of faces (\eg in the order of 20,000 faces) can better distinguish the models. We will leave this to future investigations. 

\begin{figure}[t]       
\centering 
\includegraphics[width=.18\linewidth]{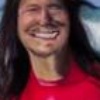}
\includegraphics[width=.18\linewidth]{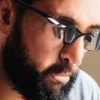}
\includegraphics[width=.18\linewidth]{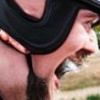}
\includegraphics[width=.18\linewidth]{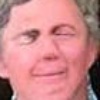}
\includegraphics[width=.18\linewidth]{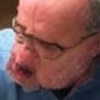}
\includegraphics[width=.18\linewidth]{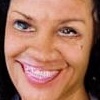}
\includegraphics[width=.18\linewidth]{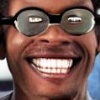}
\includegraphics[width=.18\linewidth]{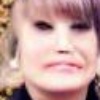}
\includegraphics[width=.18\linewidth]{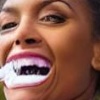}
\includegraphics[width=.18\linewidth]{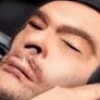}
\includegraphics[width=.18\linewidth]{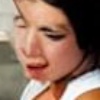}
\includegraphics[width=.18\linewidth]{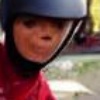}
\includegraphics[width=.18\linewidth]{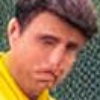}
\includegraphics[width=.18\linewidth]{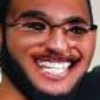}
\includegraphics[width=.18\linewidth]{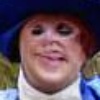}
\includegraphics[width=.18\linewidth]{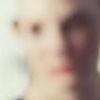}
\includegraphics[width=.18\linewidth]{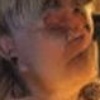}
\includegraphics[width=.18\linewidth]{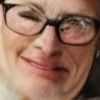}
\includegraphics[width=.18\linewidth]{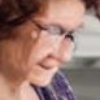}
\includegraphics[width=.18\linewidth]{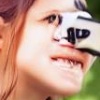}
\includegraphics[width=.18\linewidth]{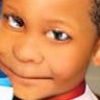}
\includegraphics[width=.18\linewidth]{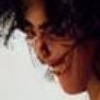}
\includegraphics[width=.18\linewidth]{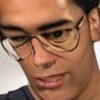}
\includegraphics[width=.18\linewidth]{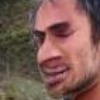}
\includegraphics[width=.18\linewidth]{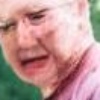}
\includegraphics[width=.18\linewidth]{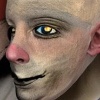}
\includegraphics[width=.18\linewidth]{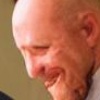}
\includegraphics[width=.18\linewidth]{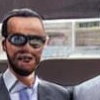}
\includegraphics[width=.18\linewidth]{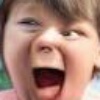}
\includegraphics[width=.18\linewidth]{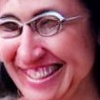}
\includegraphics[width=.18\linewidth]{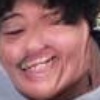}
\includegraphics[width=.18\linewidth]{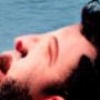}
\includegraphics[width=.18\linewidth]{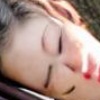}
\includegraphics[width=.18\linewidth]{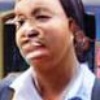}
\includegraphics[width=.18\linewidth]{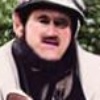}
\caption{Low quality faces generated by Stable Diffusion.}
\label{fig:bad_stable}
\end{figure}

\begin{figure}[t]       
\centering 
\includegraphics[width=.18\linewidth]{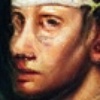}
\includegraphics[width=.18\linewidth]{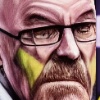}
\includegraphics[width=.18\linewidth]{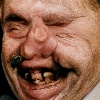}
\includegraphics[width=.18\linewidth]{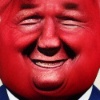}
\includegraphics[width=.18\linewidth]{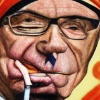}
\includegraphics[width=.18\linewidth]{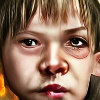}
\includegraphics[width=.18\linewidth]{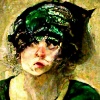}
\includegraphics[width=.18\linewidth]{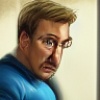}
\includegraphics[width=.18\linewidth]{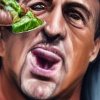}
\includegraphics[width=.18\linewidth]{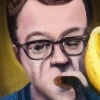}
\includegraphics[width=.18\linewidth]{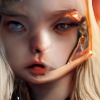}
\includegraphics[width=.18\linewidth]{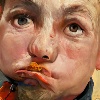}
\includegraphics[width=.18\linewidth]{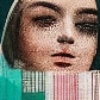}
\includegraphics[width=.18\linewidth]{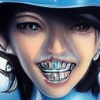}
\includegraphics[width=.18\linewidth]{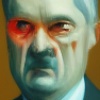}
\includegraphics[width=.18\linewidth]{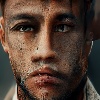}
\includegraphics[width=.18\linewidth]{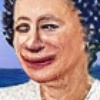}
\includegraphics[width=.18\linewidth]{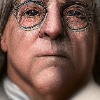}
\includegraphics[width=.18\linewidth]{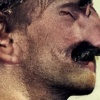}
\includegraphics[width=.18\linewidth]{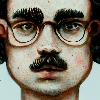}
\includegraphics[width=.18\linewidth]{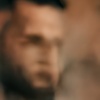}
\includegraphics[width=.18\linewidth]{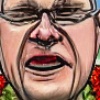}
\includegraphics[width=.18\linewidth]{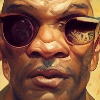}
\includegraphics[width=.18\linewidth]{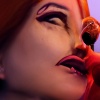}
\includegraphics[width=.18\linewidth]{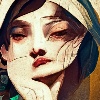}
\includegraphics[width=.18\linewidth]{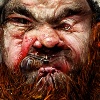}
\includegraphics[width=.18\linewidth]{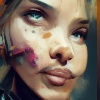}
\includegraphics[width=.18\linewidth]{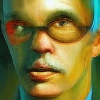}
\includegraphics[width=.18\linewidth]{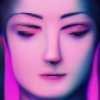}
\includegraphics[width=.18\linewidth]{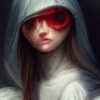}
\includegraphics[width=.18\linewidth]{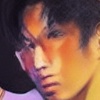}
\includegraphics[width=.18\linewidth]{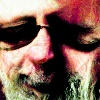}
\includegraphics[width=.18\linewidth]{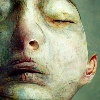}
\includegraphics[width=.18\linewidth]{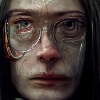}
\includegraphics[width=.18\linewidth]{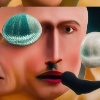}
\caption{Low quality faces generated by Midjourney.}
\label{fig:bad_mid}
\end{figure}

\begin{figure}[t]       
\centering 
\includegraphics[width=.18\linewidth]{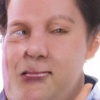}
\includegraphics[width=.18\linewidth]{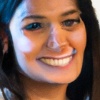}
\includegraphics[width=.18\linewidth]{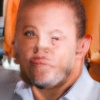}
\includegraphics[width=.18\linewidth]{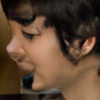}
\includegraphics[width=.18\linewidth]{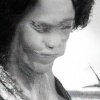}
\includegraphics[width=.18\linewidth]{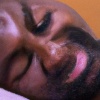}
\includegraphics[width=.18\linewidth]{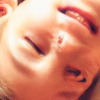}
\includegraphics[width=.18\linewidth]{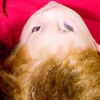}
\includegraphics[width=.18\linewidth]{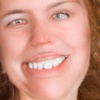}
\includegraphics[width=.18\linewidth]{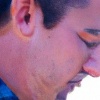}
\includegraphics[width=.18\linewidth]{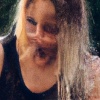}
\includegraphics[width=.18\linewidth]{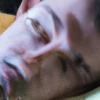}
\includegraphics[width=.18\linewidth]{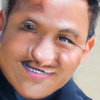}
\includegraphics[width=.18\linewidth]{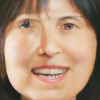}
\includegraphics[width=.18\linewidth]{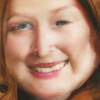}
\includegraphics[width=.18\linewidth]{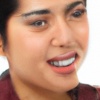}
\includegraphics[width=.18\linewidth]{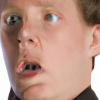}
\includegraphics[width=.18\linewidth]{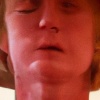}
\includegraphics[width=.18\linewidth]{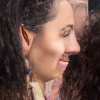}
\includegraphics[width=.18\linewidth]{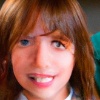}
\includegraphics[width=.18\linewidth]{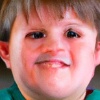}
\includegraphics[width=.18\linewidth]{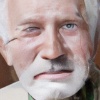}
\includegraphics[width=.18\linewidth]{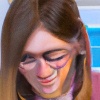}
\includegraphics[width=.18\linewidth]{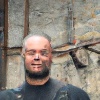}
\includegraphics[width=.18\linewidth]{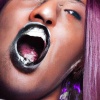}
\includegraphics[width=.18\linewidth]{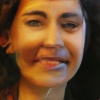}
\includegraphics[width=.18\linewidth]{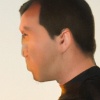}
\includegraphics[width=.18\linewidth]{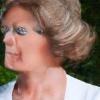}
\includegraphics[width=.18\linewidth]{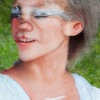}
\includegraphics[width=.18\linewidth]{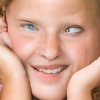}
\includegraphics[width=.18\linewidth]{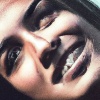}
\includegraphics[width=.18\linewidth]{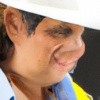}
\includegraphics[width=.18\linewidth]{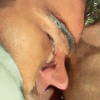}
\includegraphics[width=.18\linewidth]{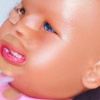}
\includegraphics[width=.18\linewidth]{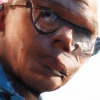}

\caption{Low quality faces generated by DALL-E 2.}
\label{fig:bad_dalle}
\end{figure}

Some low quality generated faces by models are shown in Figs.~\ref{fig:bad_stable},~\ref{fig:bad_mid}, and~\ref{fig:bad_dalle}. Models often struggle in generating eyeglasses, eyeballs, occluded faces, profile faces, and maintaining symmetry.

\section{Conclusion}

To the best of our knowledge, we are the first to evaluate the quality of generated faces in the wild. We found that Stable Diffusion generates more realistic faces. In the majority of cases, however, it is possible for a human to tell whether a face is real or generated, indicating that there is still a large gap to close. To this end, we suggest the following directions for future work in this area:

\begin{itemize}
    \item Due to limited access to DALL-E 2, we were not able to include a lot of faces from this model in our evaluation. We encourage researchers to consider larger sets of generated faces, especially from DALL-E 2 to compare and evaluate models. 
    
    \item We found that some faces look very realistic, triggering the thought that maybe they have been copied from the training set, in whole or in part. We encourage the creators of these systems to either conduct memorization tests, or release their training set for independent researchers to investigate this. We also found that some generated faces contain watermarks.
    
    \item Our data and methodology allows probing the generative models for their potential use in generating deepfakes, as well as the societal biases inherent in them. For example, annotating the faces in our dataset in terms of race, gender, and age can inform the community about the algorithmic biases in the generators.

    \item FID score uses embeddings from models trained on ImageNet. This can be problematic for evaluating the quality of faces. Classification models specifically trained on face datasets might be better options. In addition to FID, other scores such as SSIM~\cite{wang2004image}, LPIPS (learned perceptual image patch similarity)~\cite{zhang2018unreasonable}, and human judgments can be used to evaluate the generators. Further, it might be a good idea to align the faces before computing the scores.

    \item Our preliminary probing reveals that Stable Diffusion also generates better images than the other two models in terms of FID (results not shown). We also found that scaling the images (50, 100, 200, 300, 500 squares) does not change the FID score much.

    \item Future work can also investigate finer facial details such as expressions (sad, happy, angry), age (young {\emph vs. old}), and viewpoint specifications (profile {\emph vs.} frontal). Lastly, models can be compared over other interesting categories such as humans, cats, dogs, cars, and bedrooms.

\end{itemize}

{\small
\bibliographystyle{plain}
\bibliography{refs}
}

\end{document}